\documentclass{article}
\usepackage{amsmath,graphicx,array}
\usepackage{spconf}

\title {END-TO-END CONTEXTUAL ASR BASED ON POSTERIOR DISTRIBUTION ADAPTATION FOR HYBRID CTC/ATTENTION SYSTEM}
\name{Zhengyi Zhang, Pan Zhou}
\address{AI Interaction Division, Sogou Inc., Beijing, China
\\\small \texttt{zhangzhengyi@sogou-inc.com},\; \texttt{zhoupan87@gmail.com}
}

\begin{document}
\ninept
\maketitle  
\begin{abstract}
End-to-end (E2E) speech recognition architectures assemble all components of traditional speech recognition system into a single model. Although it simplifies ASR system, it introduces contextual ASR drawback: the E2E model has worse performance on utterances containing infrequent proper nouns. 
In this work, we propose to add a contextual bias attention (CBA) module to attention based encoder decoder (AED) model to improve its ability of recognizing the contextual phrases.
Specifically, CBA utilizes the context vector of source attention in decoder to attend to a specific bias embedding. Jointly learned with the basic AED parameters, CBA can tell the model when and where to bias its output probability distribution. 
At inference stage, a list of bias phrases is preloaded and we adapt the posterior distributions of both CTC and attention decoder according to the attended bias phrase of CBA. 
We evaluate the proposed method on GigaSpeech 
and achieve a consistent relative improvement on recall rate of bias phrases ranging from $15\%$ to $28\%$ compared to the baseline model.
Meanwhile, our method shows a strong anti-bias ability as the performance on general tests only degrades $1.7\% $ even 2,000 bias phrases are present.
\end{abstract}
\section{Introduction}
\label{sec:intro}

End-to-End (E2E) asr models have achieved remarkable progress in the last few years \cite{William7472621,Bahdanau7472618,chiu2018state}. One of the most popular models, hybrid CTC/attention model effectively utilizes the advantages of both architectures \cite{watanabe2017hybrid}, resulting in a comparable performance to conventional DNN/HMM ASR systems \cite{hinton2012deep}. However, end-to-end models heavily rely on the scale and distribution of the training data, which means it can hardly recognize the proper nouns rarely seen in training utterances at inference stage. Moreover, traditional n-grams language models and the corresponding WFST-based \cite{hori2013speech} decoding techniques are difficult to be incorporated into the end-to-end systems \cite{chen2019end}, which can be conveniently customized in realistic applications. Obvious performance degradation can be found in some recognition scenes such as phone call through personal AI assistant \cite{scheiner2016voice} where contact names are the main targets to be correctly recognized. Basically, contextual phrases can provide additional information about what a user may say and these information is only available at inference stage \cite{chen2019end}. Context often includes the user’s contacts, locations or collection of songs which contain rare words or proper nouns. Therefore, it is necessary to inform the system the presence of these words and encourage the model to hit the corresponding phrases in decoding.

Previous works have explored the shallow fusion methods \cite{williams2018contextual,kannan2018analysis,zhao19d_interspeech} in which the contextual language model was injected as rescoring function. The n-grams of context phrases for which the system wishes to increase likelihood are compiled into a WFST. Whenever a word boundary is reached for a specific word $w$ as decoding proceeds, the system rescore for that word with its history $w_H$ employing contextual WFST score. As biasing is applied at the end of a word, proper nouns is prone to be pruned before biasing. Data augmentation methods are also examined such as utilizing synthesized audio data with proper nouns during training \cite{sim2019personalization}. However, the synthesized data are customized to a given recognition task, lacking the generalization ability on other tasks. Another resources expended augmentation strategy tends to improve proper nouns' coverage by tagging large amount of unsupervised data \cite{zhao19d_interspeech,huang2015bidirectional}. These unsupervised utterances are supposed to be decoded by a state-of-the-art conventional ASR system and only utterances with a high confidence are kept. In \cite{pundak2018deep,alon2019contextual}, an all-neural approach for contextual biasing is proposed, where a separate bias encoder component is introduced to model the contextual phrases. Although experiment results show this approach outperforms shallow-fusion biasing on many tasks, it suffers from false triggering as the biasing influence is not controllable, resulting in poor word error rate when more bias phrases are loaded. As the end-to-end system has difficulty in generating words it rarely sees, \cite{huang2020class} proposed a transformation method which maps the rare entity words to common words via pronunciation and treats the mapped words as an alternative form to the original word during recognition. With the mapping, original words are correctly recognized.

In this work, we proposed a novel approach to address this problem based on bias attention in the hybrid CTC/attention structure. We split the bias phrases into word pieces by BPE model. Next, we employ a single lstm module as bias encoder to embed the word pieces and regard the last state of lstm output as the context embeddings of a given bias phrase. Then we send the embeddings of all the bias phrases and the context vector of source attention in the transformer decoder into a contextual bias attention (CBA) module. The bias attention module consists of two simple linear transformation on context vector and bias embeddings, which represent query and key respectively in attention technique. A scaled dot-product calculation is operated which would output a spike on the specific bias phrase when context information occur in the audio segment. We then adjust the posterior probabilities of both attention decoder and CTC outputs according to the bias attention result. During decoding process, the bias process takes place before beam searching which decreases the risk of pruning. Furthermore, the proposed system has a strong ability of anti-context on utterances we do not want to bias, which means the ASR performance would only degrade little on these utterances.

The rest of the paper is organized as follows. In Section 2, we briefly review the baseline transformer CTC/attention structure and the contextual E2E modeling technique. Our proposed CBA method is presented in Section 3 followed by experimental setup and results in Section 4. We conclude this paper with our findings and future work in Section 5.

\section{BACKGROUND}
\label{sec:background}

\subsection{Hybrid CTC/attention model}
\label{ssec:Hybrid CTC/attention model}
We use the hybrid CTC/attention ASR model as our baseline which utilizes both benefits of CTC and attention decoder during the training and decoding steps \cite{watanabe2017hybrid}. The multi-tasks learning framework is employed to improve robustness and achieve fast convergence in training stage. The monotonic alignment characteristic of CTC relieves the burden of estimating the desired alignment by attention decoder. The loss function is described as 
\begin{equation}
	L_{mtl}=\lambda \mathrm{log} p_{ctc}\{y|X\} + (1-\lambda) \mathrm{log}p_{attn}\{y|X\}     \label{eq1}
\end{equation}
where $X=\{x_1,...,x_{T_{src}}\}$ and $y=\{y_1,...,y_{T_{tgt}}\}$ denote the feature sequence and target sequence respectively. $\lambda$ is a tunable parameter, which satisfies $0\leq\lambda\leq1$.

Because the attention decoder performs decoding in label steps while CTC performs it in frame steps, the model computes the probability of each partial hypothesis using CTC and attention model respectively and incorporates both probability scores in beam searching. The attention decoder is supposed to decode from starting symbol, <$sos$> and top K units in posterior probabilities are selected to construct partial hypotheses. Then the CTC prefix probabilities can be calculated for a given partial hypothesis $h$ as:
\begin{equation}
	p_{ctc}(h,...|X)=\sum_{v}p_{ctc}(h\cdot v|X)
\end{equation}
where $v$ denotes all possible label sequences except the empty string. Then the CTC prefix scores is combined with $p_{attn}(h|X) $ like in (\ref{eq1}).
\subsection{CLAS model}
\label{ssec:CLAS model}
An effective method that integrates contextual information into the E2E modeling is called CLAS \cite{pundak2018deep}, which is an all-neural structure based on Listen, Attend and Spell (LAS) model. A list of additional bias phrases is given, denoted as $z=\{z_1,...,z_N\}$, and a bias-encoder module is designed to embed each phrase into a fixed dimensional representation $ h^z=\{h^z_0,...,h^z_N\} $. We include $h^z_0=h^z_{nobias}$ here which corresponds to the no-bias option. And then a bias attention is computed over $h^z$ where the decoder hidden state $d_t$ is utilized as attention-query like the one used in audio attention. 
Given audio and previous labels, CLAS explicitly models the probability of seeing particular phrases:
\begin{equation}
	\alpha_t^{z_i}=P(z_i|X;y_{<t})
\end{equation}
where $t$ denote the decoding step, $i$ denote the index of bias phrases.

A concatenation of audio context vector and bias context vector, $c_t=[c_t^x; c_t^z]$, is then fed into the LAS decoder as usual. By this way, the bias context information can influence the output predictions:
\begin{equation}
	P(y_t|y_{<t};X;z)=\mathrm{softmax}(\mathbf{W_s}[c_t^x; c_t^z; d_t] + b_s)
\end{equation}
And it is up to the model to determine which bias phrase might be relevant in decoding steps and the target distribution would be modified.

\section{Contextual bias attention}
\label{sec:THE PROPOSED METHO}



\subsection{Architecture}
\label{ssec:Architecture}
\begin{figure}[h]
		\centering
		\centerline{\includegraphics[width=8.5cm]{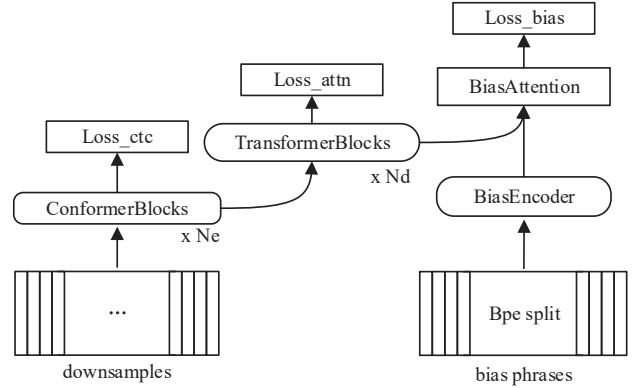}}
			\caption{Illustration of contextual bias attention system based on the conformer-transformer structure.}
			\label{fig1}
\end{figure}

The overall structure of the proposed model is shown in Fig.\ref{fig1}. We add a bias attention component compared to the standard conformer-transformer structure. The bias encoder consists of a single lstm module which use the last state of the lstm as the fixed dimensional embedding of the entire phrase. Unlike \cite{pundak2018deep}, we employ a simple scaled dot-product attention \cite{vaswani2017attention} between query and key where query denotes the context vector of source attention in the last layer of decoder and key denotes the output of the bias encoder:
\begin{equation}
\label{eq:bias-attention}
	\alpha_t^{z_i}=\mathrm{softmax}((\mathbf{W_q}c^{N_d-1}_t)(\mathbf{W_k}h^z_i)/\mathrm{sqrt}(d_{model}))
\end{equation}
where $\mathbf{W_q}$ and  $\mathbf{W_k}$ denotes a linear transformation of query and key respectively, $d_{model}$ denotes the dimension of key and $N_d$ denotes the num of layers in decoder. Bias attention is supposed to extract the potential bias phrase from a bias list in each label step according to the attention probability calculated. In our method, we generate a bias label in each label step and then the bias loss is calculated on the attention distribution. We discuss this in the next subsection.

\subsection{CBA training}
\label{ssec:Training}
In the training stage, we first need to create a bias list $z$ for each training batch. As our purpose is to improve the recognition accuracy of the proper nouns, we can naturally employ the named entity recognition annotator to label entities for reference transcripts associated with the training batch. In the following experiments, we utilize the stanford coreNLP pipeline \cite{sim2019personalization} as the extractor for NER which run several named entity recognizers and combine their results. However, only part of the reference transcripts contain named entities, and we process the rest reference transcripts by randomly selecting $n$-grams like the way used in \cite{pundak2018deep}. Therefore, a bias list $z$ will contain both the real named entities and the randomly selected word $n$-grams. Specifically, we explain the bias list creation process as follows. For a given training batch, we apply NER on the reference transcripts, $r_1,...,r_{Nbatch}$ and then split the references into two parts, one part has entity results and another one has not. As for the references without entities, $k$ word $n$-grams are randomly selected from each reference, where $k$ is selected uniformly from $[1,N\mathrm{phrases}]$ and $n$ is selected uniformly from $[1,N\mathrm{order}]$. Herein, $N\mathrm{phrases}$ denotes the maximal num of phrases can be selected in a reference and $N\mathrm{order}$ denotes the maximal order of a selected phrase.  Take a reference with 10 words as example, where $ref=[w_1,w_2,...,w_{10}]$. When we set $N\mathrm{phrases}=2$ and $N\mathrm{order}=3$, we may select 2 bias phrases, one phrase $z_1=[w_2]$ and the other one $z_2=[w_5,w_6,w_7]$ with 1 order and 3 orders respectively. The total number of bias phrases in a training batch is a variable which depends on the randomly selecting process and can differ from other batches.
Next, we need to create the bias label for each training reference which will be utilized for bias loss calculation. For a reference with bias phrases, we split the word sequence into word pieces sequence through BPE. For example, the first reference in a training batch "call hanna phone" with bias phrase "hanna" may be split into "call han @@na phone", and we label it by "0 1 1 0", where "0" means id of nobias and "1" means id of bias phrase in a bias list. Then, the second reference "I will go to shanghai" with bias phrase "shanghai" may be split into "I will go to shang @@hai", and we label it by "0 0 0 0 2 2". The rest references in the batch will be processed in the same way. The label id will indexed from "1" to the length of the bias list. After bias attention is performed at each label step, we use the attention distribution and the bias label to calculate the loss of bias attention:
\begin{equation}
	L_{bias} =-\sum_t\mathrm{log}\alpha_t^{z_t}
\end{equation}
where $z_t$ denotes the bias label at time step $t$. Furthermore, we combine the standard CTC/attention loss and the bias loss as our optimization object in the training process:
\begin{equation}
	L_{all} = -L_{mtl}+ \beta L_{bias} \label{eq:eq7}
\end{equation}
where $\beta$ denotes a tunable parameter which satisfies $0\leq\beta\leq1$.

\subsection{Posterior adaptation}
\label{ssec:Inference}
In the inference stage, we prepare the contextual phrases $z$ and send them to the bias encoder which output $h^z$ before decoding starts. During beam search, the CBA module calculate the correlation between $c^{N_d-1}_t$ and $h^z$ in each decoding step, and the biasing process is performed based on the bias phrase distribution as in  Eq. (\ref{eq:bias-attention}).
The phrase corresponding to the biggest probability index (except $h^z_0$ which means no bias index) will be biased.
We perform boosting on both posteriors of attention decoder and CTC according to the selected biased phrase. First, the biasing score $bias\_score$ will be added on the linear output corresponding to the word piece units of the attended phrase before softmax applied to attention decoder, resulting in a boosted probabilities on these word piece units. 

Next, we bias CTC posteriors corresponds to each encoded audio feature according to the source attention distribution in the last layer of attention decoder:
\begin{equation}
		\alpha_t^{h^x_i}=P(h^x_i|X;y_{<t})
\end{equation}
where $h^x$ denotes the encoded audio features and $i$ denotes the index of $h^x$. Distinguished from the bias mechanism of attention decoder, the bias score $bias\_score$ will be multiplied by the attention probability $\alpha_t^{h^x_i}$ before adding to the related word piece units of CTC linear 
output. In default, the bias score $bias\_score$ of attention decoder is consistent with the one in CTC. Consequently, both posteriors of CTC and attention decoder are biased toward the potential contextual phrases before pruning performed on the corresponding word piece units, which improves the recognition accuracy on bias phrases.

\section{EXPERIMENTS}
\label{sec:EXPERIMENTS}
\subsection{Data preparation}
We evaluate our proposed method on GigaSpeech \cite{chen2021gigaspeech}, an multi-domain English speech recognition corpus with 10,000 hours of high quality labeled audio. A variety of topics such as arts, science, sports are collected in the dataset. We apply NER of stanford coreNLP on the whole training transcripts and obtain approximate one million utterances with NER results, about one-eighth of the dataset. As for the training data without NER results, we randomly select word n-grams as described in Section \ref{ssec:Training}. We set $N_{phrases}=2$ and $N_{order}=3$ which means $1\sim2$ bias phrases each has $1\sim3$ words can be selected in a specific utterance. 

The evaluation sets in GigaSpeech consist of a dev set with 5715 utterances and a test set with 19930 utterances. We keep the dev and test set as general sets and do not apply NER to extract bias phrases. Instead, we construct bias test sets based on commonvoice training set \cite{commonvoice:2020} which is contributed by  volunteers who record their voices online and is publicly available. After all training corpus of commonvoice is labeled by NER annotator, we select the NER results by four entity classes: person names, location, organization, and other, and the filtered utterances are collected as bias test sets,denoted by PER, LOC, ORG and OTHER. A summary of the four test sets is shown in Table \ref{table1}.

In the experiments, we find some training batch with fewer bias phrases than others and we suggests adding some extra distractors to them. To examine the effect of distractors in bias training, we supplement extra n-grams when the extracted bias phrases num is less than 200. We count the words frequency in the training transcripts and exclude the top $20\%$ words and randomly select the remainder words as distractors.
\begin{table}[h]
	\caption{\textit{ Details of bias test sets. Bias list size means the num of unique bias phrases in the test set as some of utterances share the same bias phrase.}}\label{table1}
	\centering
	\begin{tabular}{|c|p{7em}<{\centering}|p{5em}<{\centering}|}
		\hline
		Test sets & Number of Utters & Bias list size \\ \hline
		PER & 16246 & 10524 \\ \hline
		LOC & 1046 & 464 \\ \hline
		ORG  & 823 & 360 \\ \hline
		OTHER & 1161 & 328 \\ \hline
	\end{tabular}

\end{table}

The total number of utterances in commonvoice training set is around 564,000 while only less than 20,000 utterances contain entities in the above four classes and "person names" occupies the highest proportion. 
\subsection{Model structure}
The model in our experiments consists of a conformer encoder with 12 blocks and a transformer decoder with 6 blocks. The full structure details can be found in the recipe of $\mathrm{"espnet/egs2/gigaspeech"}$ in espnet ASR toolkit \cite{watanabe2018espnet}. However, we use 80-dimensional filter-bank features computed on 25ms window with 10ms shift instead of the raw-wav in the espent recipe. The bias encoder is a single LSTM layer with 512 nodes. The output dimension of trainable parameter matrices in the bias attention is equal to $d_{model}$. As our system belongs to the multi-task learning framework, we set $\lambda=0.3$ in (\ref{eq1}) and $\beta=0.5$ in (\ref{eq:eq7}). During decoding, we exclude the influence of language model and only evaluate the performance of joint decoding on CTC and attention decoder.

\subsection{Bias-free testing on various test sets}
We first train the baseline hybrid CTC/attention model based on conformer-transformer structure. Then, we use the baseline model as our initial model, and add our bias encoder and bias attention components on it and carry on bias training. We compare the WER of baseline model and our bias models in decoding with an empty list of bias phrases on a variety of test sets, from general test sets of GigaSpeech to bias test sets of commonvoice. To better evaluate the biasing performance, we utilize the recall rate of bias phrases as our measurement, which means we care about the percentage of recognized bias phrases in the test set. We show the WER and recall rate in Table \ref{table2}. We find similar performance in three models when tested in bias-free mode whereas the WER on organization set degrades in bias model trained with distractors. This may be caused by the small scale of organization set as it only has 823 utterances. As for the general test sets of GigaSpeech, we only show the WER and find little performance degradation compared to baseline model.
\begin{table}
	\caption{\textit{Results of bias-free testing with compared models.}}\label{table2}
	\centering
	\begin{tabular}{|c|c|c|c|c|c|c|}
		\hline 
		& \multicolumn{2}{c|}{baseline} &\multicolumn{2}{c|}{No-distrators}
		&\multicolumn{2}{c|}{Distrators}
		\\ \cline{2-7}
		Test set & WER & recall & WER & recall& WER & recall\\ \hline
		giga\_dev & 11.8 & *** & 11.9 & ***& 11.9 & ***\\ \hline
		giga\_test & 11.7 & *** & 11.8 & ***& 11.8 & ***\\ \hline
		PER & 18.7 & 0.51 & 18.8 & 0.50& 18.7 & 0.50 \\ \hline
		LOC & 17.3 & 0.61& 17.3 & 0.61 & 17.3 & 0.61\\ \hline
		ORG & 16.0 & 0.62& 16.2 & 0.57 & 16.4 & 0.61 \\ \hline
		OTHER & 16.0 & 0.71 & 15.9 & 0.72& 16.1 & 0.71\\ \hline
		
	\end{tabular}
\end{table}

\subsection{Bias testing on bias tests}
Next, we study the performance of our bias models on bias test sets. We load the full bias list of each bias set and carry out bias testing except for the "person names" set. We divide the "person names" test set to five subsets, each loaded by the respective bias list. Because it may incur inconsistent phenomenon compared to the other three if we load all the 10524 bias phrases once in decoding. We tune the $bias\_score$ in biasing process and find similar performance, so we simply set $bias\_score=5$ in the following testing. We obtain consistent improvement on recall rate of bias phrases on three test sets, ranging from $15\%$ to $28\%$. As for the results of the five subsets of "person names", we find the recall rate improvement shrinks compared to the other three as the num of bias phrases is $2\sim3$ times larger. And the bias model with distractors training always performs better than the model without distractors training. The results are shown in Table \ref{table3}.
\begin{table}[h]
	\caption{\textit{Results of bias testing with compared models.}}\label{table3}
	\centering
	\begin{tabular}{|c|p{3em}<{\centering}|p{3em}<{\centering}|p{3em}<{\centering}|p{3em}<{\centering}|}
		\hline
		& \multicolumn{2}{c|}{No-distrators} & \multicolumn{2}{c|}{Distrators} \\ \cline{2-5}
		Test set & WER & recall & WER & recall \\ \hline
		LOC & 16.4 & 0.70 & 16.0 & 0.72 \\ \hline
		ORG & 15.4 & 0.67 & 15.2 & 0.68\\ \hline
		OTHER & 15.6 & 0.77 & 15.3 & 0.78 \\ \hline

		& \multicolumn{2}{c|}{No-distrators} & \multicolumn{2}{c|}{Distrators} \\ \cline{2-5}
		Test set & WER & recall & WER & recall \\ \hline
		PER\_sub1 & 19.7 & 0.53 & 19.0 & 0.58 \\ \hline
		PER\_sub2 & 17.8 & 0.55 & 17.4 & 0.58 \\ \hline
		PER\_sub3 & 16.8 & 0.53 & 16.2 & 0.60 \\ \hline
		PER\_sub4 & 19.9 & 0.55 & 19.3 & 0.58 \\ \hline
		PER\_sub5 & 19.4 & 0.56 & 19.0 & 0.57 \\ \hline
		PER\_aggre & 18.6 & 0.54 & 18.1 & 0.58 \\ \hline
	\end{tabular}

\end{table}
 We proceed the experiment by increasing the nums of bias phrases to 1,000 of each bias set which are selected from the "person names" bias set. Only the bias model with distractors training is examined. We achieve approximate results as the previous one, which means the bias model can adapt to a large number of phrases even the distractors are imported. The results are shown in Table \ref{table4}.
 \begin{table}[h]
  	\caption{\textit{Results of more bias distractors testing.}}\label{table4}
 	\centering
 	\begin{tabular}{|c|p{3em}<{\centering}|p{3em}<{\centering}|p{3em}<{\centering}|p{3em}<{\centering}|}
 		\hline
 		& \multicolumn{2}{c|}{Bias: full loaded} & \multicolumn{2}{c|}{Bias: 1000 loaded} \\ \cline{2-5}
 		Test set & WER & recall & WER & recall \\ \hline
 		LOC & 16.0 & 0.72 & 16.5 & 0.68 \\ \hline
 		ORG & 15.2 & 0.68 & 15.8 & 0.65 \\ \hline
 		OTHER & 15.3 & 0.78 & 15.6 & 0.76 \\ \hline

 	\end{tabular}
 \end{table}
\subsection{False triggering testing on general sets}
Finally, we show the performance of false triggering of the bias model. To prove the strong anti-context ability, we load the bias phrases into the bias encoder and perform biasing when we decode general test sets. The loaded number of bias phrases ranges from 100 to 2000 which are picked up from "person names" bias set who has the largest number of bias phrases. All the activation parameters are consistent with the previous experiments. The WER is shown in Table \ref{tabel5}.
\begin{table}[h]
	\caption{\textit{WER on $giga\_dev$ and $giga\_test$ sets with the number of bias phrases loaded. Tested on bias model with distractors training. }}\label{tabel5}
	\centering
	\begin{tabular}{|c|r|r|r|r|}
		\hline
		& \multicolumn{4}{c|}{phrase number} \\ \cline{2-5}
		Test set & 100 & 500 & 1000 & 2000 \\ \hline
		giga\_dev & 11.8 & 11.9 & 11.9 & 12.0\\ \hline
		giga\_test & 11.7 & 11.8 & 11.9 & 12.0\\ \hline
	\end{tabular}
\end{table}
Although we observe gradual degradation in WER as a function of number of bias phrases, the performance is still acceptable when loaded num increases to 2000, with only $1.7\% $ relative WER reduction in giga\_dev set.
\section{CONCLUSIONS}
In this work, we have proposed a novel contextualized E2E ASR model which leverages the contextual bias attention to guide the bias activation during beam search decoding. We described the data preparation process where NER annotator is utilized to extract bias phrases both in training and decoding. Experiments are conducted on GigaSpeech to investigate its anti-context ability and contextual bias ability. 
While the performance degrades only $1.7\%$ on general testing when 2,000 bias phrases presents,
the recall rate of bias phrases in bias testing improves consistently by  $15\%$ to $28\%$, which certifies the effectiveness, practicability of our proposal. Investigation of CBA module on other E2E architecture, e.g. RNN-T, is our next direction.

\vfill\pagebreak

%

\bibliographystyle{IEEEbib}
\bibliography{strings,refs}

\end{document}